\documentclass[journal]{IEEEtran}
\usepackage{cite}
\usepackage{amsmath,amssymb,amsfonts}
\usepackage{algorithm}
\usepackage{algpseudocode}

\makeatletter
\newcommand{\algparbox}[1]{\parbox[t]{\dimexpr\linewidth-\ALG@thistlm\relax}{#1\strut}}
\makeatother

\usepackage{graphicx}
\usepackage{booktabs}
\usepackage{multirow}
\usepackage{xcolor}
\usepackage{tikz}
\usetikzlibrary{positioning,arrows.meta,calc,fit,backgrounds}
\usepackage{pgfplots}
\pgfplotsset{compat=1.16}
\usepackage{url}

\newcommand{\method}{\textsc{PiDIn-Mixer}}
\newcommand{\methodlong}{Pipeline-Distributed Inference Mixer}
\newcommand{\methodmlp}{\textsc{PiDIn-MLP-Mixer}}
\newcommand{\methodttm}{\textsc{PiDIn-TTM}}
\newcommand{\LN}{\mathrm{LN}}
\newcommand{\R}{\mathbb{R}}
\newcommand{\sg}{\mathrm{sg}}
\newcommand{\Tok}{\mathrm{Tok}}
\newcommand{\Ch}{\mathrm{Ch}}

\begin{document}

\title{Distributed Inference of Mixers\\via Pipelined Lookup-table Encoding}

\author{Alexandros~Iosifidis%
\thanks{A. Iosifidis is with ELLIS Institute Finland, and the Data Science Research Centre, Tampere University, Finland (e-mail: alexandros.iosifidis@tuni.fi).\\The code produced for this work is available in \protect\url{https://github.com/IosifidisAlekos/pidinmixer}.}
}

\maketitle

\begin{abstract}
Pipeline-parallel inference distributes the layers of a neural network over co-located edge devices, and every split point transmits the complete or compressed intermediate activation. In this paper, we introduce a framework for pipeline-distributed inference of Mixer models that transmits one short integer code per token. Motivated by table-lookup inference, the proposed \method{} places every split point between the token-mixing and the channel-mixing sublayers of a Mixer layer. The sending device computes the token mixing and encodes every row of its output with a learned linear projection and fixed per-dimension thresholds. For short codes, the receiving device reads the channel-mixing output of every row from a table indexed by its code, which also absorbs the computations of a nonlinear decoder. For longer codes, for which the table would not fit in the memory of the receiving device, it decodes every row and computes the channel-mixing sublayer as an ordinary layer. In our experiments with MLP-Mixer-B/16 on image classification tasks and with Tiny Time Mixers on time-series forecasting tasks, we show that \method{} provides a trade-off between compression and accuracy. With two devices, codes of $48$ bits per token, which transmit $512\times$ fewer bits than FP32 activations for MLP-Mixer-B/16, reduce the accuracy by at most $1.55\%$ and increase the forecasting error by at most $2.1\%$, and stronger compression leads to gradually larger losses.
\end{abstract}

\begin{IEEEkeywords}
Distributed inference, pipeline parallelism, edge computing, communication-efficient machine learning, lookup tables, vector quantization, MLP-Mixer, time-series forecasting.
\end{IEEEkeywords}

\section{Introduction}
\IEEEPARstart{D}{eep} neural networks achieve strong performance in vision and time-series analysis, and many of their applications require inference close to the data, at the network edge, because of latency, privacy, energy, or connectivity constraints~\cite{zhang2025edgeshard,ye2025jupiter,peccia2024review}. When a single edge device cannot host or run a model alone, the model is distributed over several co-located devices whose aggregate memory and compute cover its demands.
Three parallelism approaches serve this purpose, namely tensor parallelism~\cite{shoeybi2019megatron}, pipeline parallelism~\cite{huang2019gpipe}, and sequence parallelism~\cite{li2023sequence}. These approaches are commonly used in accelerator clusters with high-bandwidth interconnects, where inter-device communication can become a main bottleneck~\cite{pope2023scaling,anthony2024demystifying}, and its cost increases further when they are applied to inference over bandwidth-constrained links~\cite{wei2024detransformer,liu2026astra}.

In pipeline parallelism, contiguous groups of layers are assigned to pipeline stages, and the hidden state is passed from one stage to the next~\cite{huang2019gpipe}. When this structure is used for inference over edge devices, every transition from one device to the next transmits the full hidden state of the model. For a Mixer model whose hidden state has $N$ tokens and $D$ channels, this corresponds to $ND$ real values per input and split point, for instance $150{,}528$ values (or $4{,}816{,}896$ bits in FP32) for MLP-Mixer-B/16~\cite{tolstikhin2021mixer}. Existing solutions compress the transmitted feature with learned bottlenecks, as in split computing~\cite{matsubara2022split,shao2020bottlenet,li2022attention}, transmit vector-quantization indices between sequence-parallel partitions~\cite{liu2026astra,qazi2026trace}, or quantize every activation to fewer bits~\cite{dettmers2022llmint8} which still transmits $ND$ values.

This work is motivated by table-lookup inference. LUT-NN~\cite{tang2023lutnn} learns typical inputs (centroids) for each operator of a network, precomputes the operator's results for these centroids and stores them in lookup tables. At inference, the results of the centroids closest to the actual inputs are read from the tables instead of being computed. The underlying idea goes back to the table-based inner products of product quantization~\cite{jegou2011pq}. \textsc{Maddness} uses learned hashing to select precomputed table entries so that, when one matrix is known in advance, a matrix product needs no multiply-adds~\cite{blalock2021maddness}. In all these methods, the input of an operator is represented by few integer indices and the operator's output is recovered from tables. In a distributed setting, the same principle suggests transmitting indices instead of activations, and letting the receiving device recover what it needs from tables.

Mixer architectures, namely MLP-Mixer~\cite{tolstikhin2021mixer} and the TSMixer layers from which Tiny Time Mixers (TTM) are built~\cite{ekambaram2023tsmixer,ekambaram2024ttm}, are a particularly good fit for this principle. Each of their layers consists of an MLP applied to single columns (token mixing) followed by an MLP applied to single rows (channel mixing) of the hidden state. Once a row is represented by one code, everything that the channel-mixing sublayer computes for that row is a function of the code and can be precomputed. However, using this property for pipeline-distributed inference raises three challenges. First, the codes need to be cheap to compute on the sending device. Assigning a vector to the nearest of $K$ centroids, as in LUT-NN, requires about $KD$ multiply-adds per vector, and a high accuracy requires $K$ to be large, so that the sending device would spend on encoding what it saves on transmission. Second, token mixing combines all tokens of a column, and therefore its output cannot be tabulated per row code, i.e., the codes need to be assigned where the computation becomes row-wise. Third, the receiving device continues with the layers of a trained model, so that the decoded rows need to be close to what these layers expect, and the conversion needs to preserve the accuracy that the trained model reached.

In this paper, we address these challenges by introducing \methodlong{} (\method{}). In \method{}, every split point is placed between the two sublayers of a Mixer layer, i.e., the sending device computes the token mixing of the layer and encodes every row of the token-mixed table. The codes are obtained without a search by projecting each row to a $d$-dimensional space ($d \ll D$) with a learned linear map and comparing each dimension with fixed thresholds, so that the code is an integer formed by the levels of the $d$ dimensions, as in finite scalar quantization~\cite{mentzer2024fsq} and lookup-free quantization~\cite{yu2024magvit2}. The receiving device reads the channel-mixing output of every row from a table indexed by its code. Since the table is computed once for all codes, it also absorbs a nonlinear decoder that maps the codes back to rows, without adding computations at inference. Codes that are longer than the table allows are decoded and processed by ordinary layers. A trained Mixer model is transformed to its \method{} version in two steps, i.e., the code is first learned with the pre-trained network frozen and then fine-tuned on the task together with the entire network by knowledge distillation from the original model~\cite{hinton2014distilling,romero2015fitnets}. We make the following contributions:
\begin{itemize}
    \item We propose to place the split points of a pipeline between the two sublayers of a Mixer layer, so that the sending device transmits one integer code of $b$ bits per token, i.e., $Nb$ bits per input and split point independently of the model width $D$, and the receiving device obtains the channel-mixing output of every token by a table lookup.

    \item We propose search-free codes that require $dD$ multiply-adds per token on the sending device, together with a nonlinear decoder that is absorbed by the table of the receiving device. We show that transmitting the codes and reading the table adds no approximation to the converted model.

    \item We propose a two-step conversion of trained models, which first fits the code to the frozen network and then fine-tunes both by distillation.

    \item We conduct experiments on image classification with MLP-Mixer-B/16 and on time-series forecasting with TTM, split over two and three devices, with codes of $12$ to $120$ bits per token, and we compare \method{} with the transmission of FP16, INT8 and INT4 activations, with and without fine-tuning.
\end{itemize}

The remainder of this paper is organized as follows. Section~\ref{sec:related} reviews related work, Section~\ref{sec:method} describes \method{}, Section~\ref{sec:experiments} provides experimental results, and Section~\ref{sec:conclusion} concludes the paper.

\section{Background \& Related Work}\label{sec:related}

\subsection{Mixer Architectures}
MLP-Mixer~\cite{tolstikhin2021mixer} maps an image to $N$ patches (tokens), embeds each patch into $D$ channels, and arranges the result as $\mathbf{X}\in\R^{N\times D}$. Each layer applies two residual sublayers:
\begin{align}
\mathbf{U}_{:,c} &= \mathbf{X}_{:,c} + \Tok\big(\LN_1(\mathbf{X})_{:,c}\big), \quad c=1,\dots,D, \label{eq:tok}\\
\mathbf{Y}_{t,:} &= \mathbf{U}_{t,:} + \Ch\big(\LN_2(\mathbf{U}_{t,:})\big), \quad t=1,\dots,N, \label{eq:ch}
\end{align}
where $\Tok(\mathbf{v})=\mathbf{W}_2\phi(\mathbf{W}_1\mathbf{v}+\mathbf{b}_1)+\mathbf{b}_2$ is an MLP with hidden width $D_{\mathrm{tok}}$ applied to every column (token mixing), $\Ch(\cdot)$ is an MLP with hidden width $D_{\mathrm{ch}}$ applied to every row (channel mixing), $\LN_1(\cdot),\LN_2(\cdot)$ are LayerNorm layers, and $\phi(\cdot)$ is the GELU activation function. After the last layer, the rows are normalized, averaged and mapped to class scores by a linear head.
TSMixer~\cite{ekambaram2023tsmixer} and Tiny Time Mixers (TTM)~\cite{ekambaram2024ttm} use a related two-sublayer structure for time series. Every channel of a series is cut into patches and processed independently as a patches-by-features table, with a gated MLP over the patches (patch mixing, the counterpart of token mixing) and a gated MLP over the features (feature mixing, the counterpart of channel mixing). In TTM, the backbone also includes levels between which the table is reshaped to fewer patches with more features (adaptive patching)~\cite{ekambaram2024ttm,granitetsfm}. A linear adapter connects the backbone to a small decoder made of the same layers and to a linear forecast head.
Because $\LN_2(\cdot)$ and $\Ch(\cdot)$ (or the corresponding feature mixing) act on one row at a time, $\mathbf{Y}_{t,:}$ is a function of $\mathbf{U}_{t,:}$ alone. This is the property that \method{} exploits.

\subsection{Distributed Inference at the Edge}
Tensor parallelism partitions weight matrices and exchanges partial results after every sublayer~\cite{shoeybi2019megatron}. Pipeline parallelism assigns contiguous layers to devices and passes the full hidden state at every stage transition~\cite{huang2019gpipe}, and sequence parallelism partitions the token sequence across devices~\cite{li2023sequence}. When the bandwidth is limited, the exchanged data can dominate inference latency~\cite{pope2023scaling,wei2024detransformer,liu2026astra}, and edge-oriented systems address device selection and model partitioning~\cite{zhang2025edgeshard,ye2025jupiter}. A review of such methods can be found in~\cite{peccia2024review}.

Compression of the exchanged data has been studied in two settings. In split computing, a network is divided into a head on the edge device and a tail on an edge server, and the intermediate features are compressed before transmission~\cite{matsubara2022split}, e.g., by a learned encoder--decoder pair~\cite{shao2020bottlenet} or by attention-based feature selection~\cite{li2022attention}. Dynamic split computing selects the split point at run time among the natural bottlenecks of a network, i.e., the layers whose output is smaller than the input of the network, according to the current data rate and batch size~\cite{bakhtiarnia2023dynamic}. In sequence-parallel Transformer inference, \textsc{Voltage} transmits uncompressed hidden-state partitions~\cite{hu2024voltage}, \textsc{Prism} transmits segment means~\cite{qazi2025prism}, \textsc{Astra} transmits low-bit vector-quantized representations of non-local tokens, which are used in a mixed-precision attention mechanism~\cite{liu2026astra}, and \textsc{Trace} transmits vector quantization indices of the post-attention residual together with precomputed key/value buffers that remove the projections of the receiving device~\cite{qazi2026trace}. \method{} also transmits integer codes and replaces computations of the receiving device by precomputed tables. However, it targets pipeline rather than sequence parallelism, and it exploits the row-wise structure of the channel-mixing sublayer of Mixers, so that the receiving device reads the complete output of a sublayer rather than a projection.

\subsection{Quantization and Table-Lookup Building Blocks}
\textit{Vector quantization} represents a vector by the index of its nearest codeword~\cite{gersho1991vq}. VQ-VAE trains codebooks inside neural networks with a straight-through gradient and a commitment loss~\cite{vandenoord2017vqvae}, where the straight-through estimator copies the gradient through a non-differentiable operation~\cite{bengio2013ste}. Finding the nearest of $K$ codewords requires $KD$ multiply-adds per vector.
\textit{Finite scalar quantization} projects a representation to a few dimensions and rounds each dimension to a small set of values, so that the codebook is the implicit product of these sets and needs no search~\cite{mentzer2024fsq}. Lookup-free quantization uses binary values per dimension, so that no codebook lookup is needed~\cite{yu2024magvit2}.
Product quantization splits vectors into sub-vectors and estimates inner products by adding precomputed table entries~\cite{jegou2011pq}. \textsc{Maddness} approximates matrix products by learned hashing and precomputed tables~\cite{blalock2021maddness}, and LUT-NN learns centroids for every operator and replaces linear layers by table lookups~\cite{tang2023lutnn}.
\textit{Activation quantization} with an absolute-maximum scale per vector (vector-wise quantization) is used for 8-bit inference of Transformers~\cite{dettmers2022llmint8}, and quantization-aware training simulates quantization during training to preserve accuracy~\cite{jacob2018quantization}. Knowledge distillation trains a student model on the softened outputs of a teacher~\cite{hinton2014distilling}, and FitNets additionally match intermediate representations~\cite{romero2015fitnets}. 
\method{} combines these elements, i.e., finite scalar quantization-style codes for search-free coding, LUT-NN-style tables that replace a whole sublayer on the receiving device, and straight-through training with knowledge distillation. Activation quantization with and without quantization-aware fine-tuning serves as our baseline.

\section{Methodology}\label{sec:method}
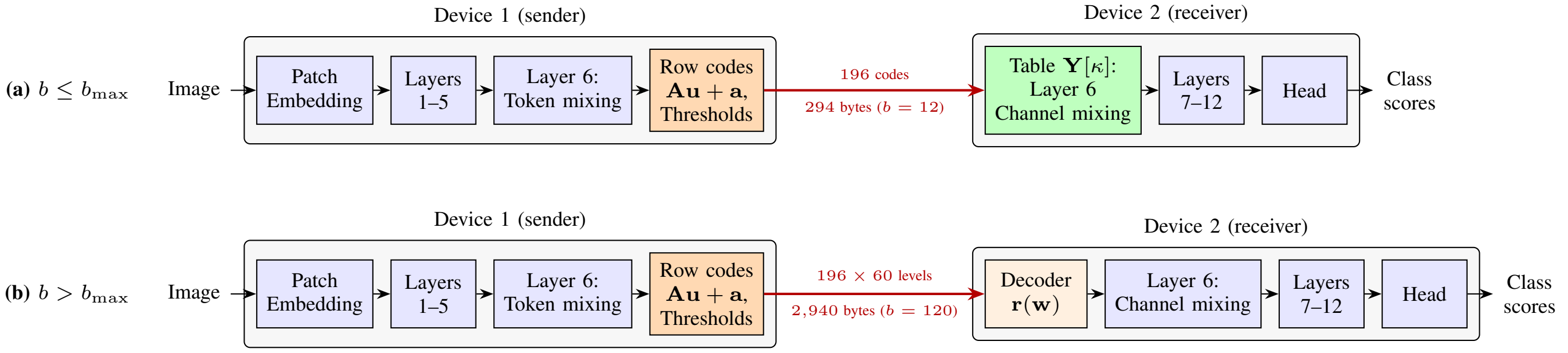
\begin{figure*}[t]
\centering
\resizebox{\textwidth}{!}{%
\begin{tikzpicture}[
  dev/.style={draw, rounded corners=2pt, fill=gray!6, inner sep=4pt},
  dense/.style={draw, fill=blue!10, minimum width=10mm, minimum height=8mm, font=\scriptsize, align=center},
  enc/.style={draw, fill=orange!30, minimum width=12mm, minimum height=8mm, font=\scriptsize, align=center},
  tab/.style={draw, fill=green!25, minimum width=12mm, minimum height=8mm, font=\scriptsize, align=center},
  dec/.style={draw, fill=orange!12, minimum width=12mm, minimum height=8mm, font=\scriptsize, align=center},
  io/.style={font=\scriptsize, align=center},
  case/.style={font=\scriptsize\bfseries, anchor=east, align=right},
  ar/.style={-{Stealth[length=1.6mm]}, thin},
  msg/.style={-{Stealth[length=2mm]}, thick, red!70!black}]
\node[io] (img) at (0,0) {Image};
\node[case] at ([xshift=-2mm]img.west) {(a) $b\le b_{\max}$};
\node[dense, right=3mm of img] (e1) {Patch\\Embedding};
\node[dense, right=2mm of e1] (l1) {Layers\\1--5};
\node[dense, right=2mm of l1] (t6) {Layer 6:\\Token mixing};
\node[enc, right=2mm of t6] (c6) {Row codes\\$\mathbf{A}\mathbf{u}+\mathbf{a}$,\\Thresholds};
\begin{scope}[on background layer]\node[dev, fit=(e1)(c6), label={[font=\scriptsize]above:Device 1 (sender)}] {};\end{scope}
\node[tab, right=26mm of c6] (y6) {Table $\mathbf{Y}[\kappa]$:\\Layer 6\\Channel mixing};
\node[dense, right=2mm of y6] (l7) {Layers\\7--12};
\node[dense, right=2mm of l7] (hd) {Head};
\begin{scope}[on background layer]\node[dev, fit=(y6)(hd), label={[font=\scriptsize]above:Device 2 (receiver)}] {};\end{scope}
\node[io, right=3mm of hd] (out) {Class\\scores};
\draw[ar] (img)--(e1); \draw[ar] (e1)--(l1); \draw[ar] (l1)--(t6); \draw[ar] (t6)--(c6);
\draw[msg] (c6)--node[above, font=\tiny, text=red!70!black]{$196$ codes}node[below, font=\tiny, text=red!70!black]{$294$ bytes ($b=12$)}(y6);
\draw[ar] (y6)--(l7); \draw[ar] (l7)--(hd); \draw[ar] (hd)--(out);
\node[io] (img2) at (0,-2.4) {Image};
\node[case] at ([xshift=-2mm]img2.west) {(b) $b>b_{\max}$};
\node[dense, right=3mm of img2] (e2) {Patch\\Embedding};
\node[dense, right=2mm of e2] (l2) {Layers\\1--5};
\node[dense, right=2mm of l2] (t62) {Layer 6:\\Token mixing};
\node[enc, right=2mm of t62] (c62) {Row codes\\$\mathbf{A}\mathbf{u}+\mathbf{a}$,\\Thresholds};
\begin{scope}[on background layer]\node[dev, fit=(e2)(c62), label={[font=\scriptsize]above:Device 1 (sender)}] {};\end{scope}
\node[dec, right=26mm of c62] (d62) {Decoder\\$\mathbf{r}(\mathbf{w})$};
\node[dense, right=2mm of d62] (ch62) {Layer 6:\\Channel mixing};
\node[dense, right=2mm of ch62] (l72) {Layers\\7--12};
\node[dense, right=2mm of l72] (hd2) {Head};
\begin{scope}[on background layer]\node[dev, fit=(d62)(hd2), label={[font=\scriptsize]above:Device 2 (receiver)}] {};\end{scope}
\node[io, right=3mm of hd2] (out2) {Class\\scores};
\draw[ar] (img2)--(e2); \draw[ar] (e2)--(l2); \draw[ar] (l2)--(t62); \draw[ar] (t62)--(c62);
\draw[msg] (c62)--node[above, font=\tiny, text=red!70!black]{$196\times60$ levels}node[below, font=\tiny, text=red!70!black]{$2{,}940$ bytes ($b=120$)}(d62);
\draw[ar] (d62)--(ch62); \draw[ar] (ch62)--(l72); \draw[ar] (l72)--(hd2); \draw[ar] (hd2)--(out2);
\end{tikzpicture}}
\caption{\textbf{\method{} overview} for MLP-Mixer-B/16 on $P=2$ devices. The split layer (layer 6) is divided between its sublayers, and the sending device computes the token mixing and encodes every row of the result. (a) For codes of up to $b_{\max}$ bits, the sending device transmits one code per token ($196$ codes of $12$ bits for $K=4{,}096$), and the receiving device reads the channel-mixing output of every row from a table indexed by its code. (b) For longer codes, the sending device transmits the $d$ levels of every token ($196\times60$ levels of $2$ bits for $b=120$), and the receiving device rebuilds every row with the decoder and computes the channel-mixing sublayer as an ordinary layer. In both cases, the receiving device then continues with the ordinary layers.}
\label{fig:overview}
\end{figure*}

\method{} targets pipeline-distributed inference of a Mixer model with $L$ layers over $P$ co-located devices. The layers are split into $P$ contiguous ranges, with device $p$ holding layers $[s_p,e_p)$. The first device also holds the patch embedding, and the last device holds the final normalization and the classification or forecasting/regression head. The last layer of every sending device is the \emph{split layer}, i.e., its token-mixing sublayer runs on the sending device, and its channel-mixing sublayer runs on the receiving device. Depending on the length $b$ of the codes, the receiving device either reads the output of the channel-mixing sublayer from a table, for $b\le b_{\max}$ (as described in Section~\ref{sec:split}), or computes it as an ordinary layer after a decoder, for $b>b_{\max}$ (as described in Section~\ref{sec:large}). Fig.~\ref{fig:overview} shows the two cases for $P=2$, and Table~\ref{tab:notation} summarizes our notation.

\subsection{Model Conversion and Inference}\label{sec:overview}
\method{} converts a trained Mixer model in three steps. First, the original model is trained on the target task in the usual way, e.g., from random initialization or by fine-tuning pre-trained weights until convergence. Second, the code of every split layer is calibrated on the training data. Third, the model is fine-tuned in two steps, i.e., the code alone with the original network frozen, followed by the code and the network together. For $b\le b_{\max}$, a fourth step is applied where the decoder and the channel-mixing sublayer of every split layer are evaluated once for every code, and the results are stored on the receiving device.

At inference, the split point between device $p$ and device $p+1$ involves five steps per input:
\begin{enumerate}
\item Device $p$ runs its ordinary layers on its input and the token-mixing sublayer of its split layer, which gives the token-mixed table $\mathbf{U}\in\R^{N\times D}$.
\item Device $p$ projects every row of $\mathbf{U}$ to $d$ numbers and compares each of them with fixed thresholds, which gives one integer code per token.
\item Device $p$ sends the $N$ codes, of $b$ bits each (for TTM, together with the mean and the standard deviation of each series) to device $p+1$.
\item For $b\le b_{\max}$, device $p+1$ reads the channel-mixing output of every row from its table, thus, it performs no multiplication for the split layer. For $b>b_{\max}$, it rebuilds every row with the decoder and computes the channel-mixing sublayer as an ordinary layer. 
\item Device $p+1$ continues with its ordinary layers.
\end{enumerate}

\subsection{Splitting a Mixer Layer Between Its Sublayers}\label{sec:split}
Let layer $e_p$ be the last layer of device $p<P$, with input $\mathbf{X}$. The sending device computes the token-mixed table $\mathbf{U}$ of this layer using Eq.~\eqref{eq:tok}, and encodes every row $\mathbf{u}_t=\mathbf{U}_{t,:}$ as described in Section~\ref{sec:codes}. By Eq.~\eqref{eq:ch}, the output row $\mathbf{Y}_{t,:}$ of the layer is a function of $\mathbf{u}_t$ alone, so that the receiving device can obtain it from the code of $\mathbf{u}_t$ by a table lookup, as described in Section~\ref{sec:table}. With this split point, the token mixing combining all tokens is computed exactly, and the channel-mixing sublayer, which accounts for most of the layer's computation ($88.9\%$ of the multiply-adds of an MLP-Mixer-B/16 layer) is replaced by one table lookup per token. 

The table stores $K\times D$ values, so that its size doubles with every additional bit per token (e.g., $12.6$~MB for $12$ bits and $201$~MB for $16$ bits for MLP-Mixer-B/16). Depending on the memory available at the receiving device, the table can therefore be used only up to a maximum code length $b_{\max}$. For longer codes, the sending device transmits the $d$ levels of every token, and the receiving device rebuilds each row with the decoder and computes the channel-mixing sublayer as in an ordinary layer, as described in Section~\ref{sec:large}.

\begin{table}[t!]
\caption{Summary of notation.}
\label{tab:notation}
\centering
\footnotesize
\begin{tabular}{@{}ll@{}}
\toprule
Symbol & Description\\
\midrule
$P$, $p$ & Number of devices, and device index\\
$N$, $D$ & Tokens (patches), and channels of a layer\\
$\mathbf{X},\mathbf{U},\mathbf{Y}$ & Split-layer input, token-mixed table, and output\\
$\LN_1(\cdot),\LN_2(\cdot)$ & Trained normalizations of a layer\\
$\Tok(\cdot),\Ch(\cdot)$ & Trained MLPs of a layer\\
$d$, $L$, $K=L^d$ & Projected dimensions, levels per dimension, and codes\\
$b=\log_2K$ & Bits per token, i.e., $d\log_2L$\\
$b_{\max}$ & Maximum code length for which the table is used\\
$\mathbf{A},\mathbf{a}$ & Projection of a row to $d$ numbers (sending device)\\
$\tau_m$, $\omega_\ell$ & Thresholds, and value of level $\ell$\\
$\ell_{t,i}$, $\kappa_t$ & Level of dimension $i$ of row $t$, and code of row $t$\\
$\theta_i$ & Learned surrogate temperature of dimension $i$\\
$\mathbf{B},\mathbf{b},g(\cdot)$ & Linear decoder, and decoder MLP\\
$\mathbf{r}(\kappa)$, $\mathbf{Y}[\kappa]$ & Decoded row, and table entry (channel-mixing output)\\
\bottomrule
\end{tabular}
\end{table}

\subsection{Row Codes}\label{sec:codes}
Every row is projected to $d$ numbers, and each number is compared with fixed thresholds, as in finite scalar quantization~\cite{mentzer2024fsq}:
\begin{align}
\mathbf{y}_t&=\mathbf{A}\mathbf{u}_t+\mathbf{a} \quad \in\R^d, \label{eq:proj}\\
\ell_{t,i}&=\textstyle\sum_{m=1}^{L-1}\mathbb{I}[y_{t,i}>\tau_m],\quad i=1,\dots,d, \label{eq:levels}\\
\kappa_t&=\textstyle\sum_{i=1}^{d}\ell_{t,i}L^{\,i-1}, \label{eq:index}
\end{align}
where $\tau_m=\mathrm{logit}\big((m-\tfrac12)/(L-1)\big)$, i.e., $-1.61$, $0$ and $1.61$ for $L=4$. Thus, the codebook contains $K=L^d$ codes, and every code takes $b=\log_2K=d\log_2L$ bits, where $L$ is a power of two. The projection $\mathbf{A}$ is initialized from the leading $d$ principal components of rows collected from training inputs, scaled so that every projected dimension has a standard deviation of $1.5$. 
The code of a row is obtained with $dD$ multiply-adds and $d(L-1)$ comparisons. Thus, every token is represented by one integer code $\kappa_t$, which is the only value transmitted for that token. Since $\kappa_t$ determines the levels $\ell_{t,1},\dots,\ell_{t,d}$ uniquely, they can be recovered from it. Level $\ell$ of every dimension has the value $\omega_\ell=2\ell/(L-1)-1$, so that the code of row $t$ corresponds to the vector $\mathbf{w}_t=(\omega_{\ell_{t,1}},\dots,\omega_{\ell_{t,d}})\in\R^d$, which is the input of the decoder. For the $N$ tokens, the decoder is applied to $\mathbf{w}_1,\dots,\mathbf{w}_N$, i.e., to every row separately.

Since thresholding has no useful gradient, the forward pass during training uses the levels of Eq.~\eqref{eq:levels}, and the backward pass uses a soft count:
\begin{equation}
s_{t,i}=\textstyle\sum_{m=1}^{L-1}\sigma\bigg((y_{t,i}-\tau_m)/\theta_i\bigg),\quad \tilde\ell_{t,i}=s_{t,i}+\sg(\ell_{t,i}-s_{t,i}),
\end{equation}
where $\sigma(x)=1/(1+e^{-x})$ is the logistic sigmoid function, so that the gradients pass straight through the thresholds~\cite{bengio2013ste}, and $\sg(\cdot)$ is the stop-gradient operator. LUT-NN reports that learning the softmax temperature of its soft assignment improves the accuracy substantially~\cite{tang2023lutnn}. Following this, we learn the temperature $\theta_i=e^{\vartheta_i}$ of every dimension, which is initialized to half the spacing between the thresholds.

\subsection{Decoder and Table of the Receiving Device}\label{sec:table}
A code is mapped back to a row by a linear decoder and an MLP correction:
\begin{equation}
\mathbf{r}(\mathbf{w})=\mathbf{B}\mathbf{w}+\mathbf{b}+g(\mathbf{w}),\qquad g(\mathbf{w})=\mathbf{W}_4\,\phi(\mathbf{W}_3\mathbf{w}+\mathbf{b}_3)+\mathbf{b}_4,
\label{eq:decoder}
\end{equation}
with $\mathbf{B}$ being initialized as the reconstruction corresponding to the projection $\mathbf{A}$. $\mathbf{W}_3,\mathbf{b}_3$ are initialized with values drawn uniformly from $[-1/\sqrt{d},1/\sqrt{d}]$, and $\mathbf{W}_4,\mathbf{b}_4$ are initialized to zeros. Thus, $g(\cdot)$ is zero at the start of training, so that training starts from the calibrated linear code.

During training, the split layer outputs $\hat{\mathbf{Y}}_{t,:}=\mathbf{r}(\mathbf{w}_t)+\Ch(\LN_2(\mathbf{r}(\mathbf{w}_t)))$. After training, the receiving device stores, for every code $\kappa$ with levels $\boldsymbol{\ell}(\kappa)$ and values $\mathbf{w}(\kappa)$, the table
\begin{equation}
\mathbf{Y}[\kappa]=\mathbf{r}\big(\mathbf{w}(\kappa)\big)+\Ch\bigg(\LN_2\big(\mathbf{r}(\mathbf{w}(\kappa))\big)\bigg),\quad \kappa=0,\dots,K-1,\label{eq:table}
\end{equation}
i.e., $K\times D$ values. For MLP-Mixer-B/16 with $K=4{,}096$, this corresponds to $12.6$~MB. The decoder MLP $g(\cdot)$ is evaluated only to fill the table, and therefore it does not add computations at inference. The token-mixing MLP on the sending device, on the other hand, runs for every inference since its inputs are real-valued.

Transmitting the codes and reading the table does not add any approximation to the one that is already present in the converted model, i.e., the model trained with the codes in its split layers. The sending device computes the projection $\mathbf{y}_t$ and the levels $\ell_{t,i}$ of every row in the same way as the converted model. Since the code $\kappa_t$ of Eq.~\eqref{eq:index} determines these levels uniquely, the receiving device recovers from it the levels and, therefore, the values $\mathbf{w}_t$. The table entry $\mathbf{Y}[\kappa_t]$ of Eq.~\eqref{eq:table} is then obtained by applying the decoder and the channel-mixing sublayer to $\mathbf{w}_t$, which is the same computation that gives the output $\hat{\mathbf{Y}}_{t,:}$ of the split layer in the converted model. Thus, the split model computes the same output as the converted model.

\subsection{Codes Longer Than the Table}\label{sec:large}
For codes longer than the maximum code length of the table, i.e., for $b>b_{\max}$, the sending device transmits the $d$ levels of every token, i.e., $d$ values of $b = \log_2L$ bits in total, and the receiving device computes $\mathbf{r}(\mathbf{w}_t)$ and the channel-mixing sublayer as ordinary layers. The split model still computes the same output as the converted model. Compared with the original model split over the same devices, the channel-mixing sublayer of the split layer is computed in both cases, i.e., it only moves from the sending to the receiving device, and the only additional computations are those of the projection and of the decoder (for MLP-Mixer-B/16 with $d=60$ this corresponds to $1.13\%$ of the model computations). Thus, codes longer than the table do not reduce the number of computations, but they lead to a higher accuracy at the cost of a larger data transfer, as seen in Section~\ref{sec:curve}.

\subsection{Two-Step Training}\label{sec:training}
In the first step, all weights of the original model are frozen, and the encoding parameters $\{\mathbf{A},\mathbf{a},\mathbf{B},\mathbf{b},g,\vartheta\}$ of every split layer are trained using the loss
\begin{equation}
\mathcal{L}_1=\sum_{\text{split layers}}\frac{\Vert\hat{\mathbf{Y}}-\mathbf{Y}^{\circ}\Vert_F^2}{\Vert\mathbf{Y}^{\circ}\Vert_F^2}+\mathcal{L}_{\mathrm{out}}\big(\hat{\mathbf{z}},\mathbf{z}^{\circ}\big).
\label{eq:step1}
\end{equation}
The first term compares the output $\hat{\mathbf{Y}}$ of every split layer with the output $\mathbf{Y}^{\circ}$ of the corresponding layer of the original model for the same input, relative to the energy of $\mathbf{Y}^{\circ}$, i.e., it trains the code to reproduce the intermediate representations of the original model~\cite{romero2015fitnets}. The second term compares the output $\hat{\mathbf{z}}$ of the converted model with the output $\mathbf{z}^{\circ}$ of the original model. For classification, $\hat{\mathbf{z}}$ and $\mathbf{z}^{\circ}$ are the class scores and $\mathcal{L}_{\mathrm{out}}(\cdot)$ is the distillation loss~\cite{hinton2014distilling}
\begin{equation}
\mathcal{L}_{\mathrm{out}}\big(\hat{\mathbf{z}},\mathbf{z}^{\circ}\big)=T^2\sum_{j=1}^{J}p_j^{\circ}\log\frac{p_j^{\circ}}{\hat{p}_j},
\label{eq:kd}
\end{equation}
where $\hat{\mathbf{p}}=\mathrm{softmax}(\hat{\mathbf{z}}/T)$ and $\mathbf{p}^{\circ}=\mathrm{softmax}(\mathbf{z}^{\circ}/T)$ are the class probabilities of the two models at temperature $T$, and $J$ is the number of classes. For forecasting/regression, $\hat{\mathbf{z}},\mathbf{z}^{\circ}\in\R^{M}$ are the $M$ predicted values and $\mathcal{L}_{\mathrm{out}}(\cdot)$ is their mean squared difference
\begin{equation}
\mathcal{L}_{\mathrm{out}}\big(\hat{\mathbf{z}},\mathbf{z}^{\circ}\big)=\frac{1}{M}\big\Vert\hat{\mathbf{z}}-\mathbf{z}^{\circ}\big\Vert_2^2.
\label{eq:mse}
\end{equation}

In the second step, all parameters are trained with the task loss and the same output distillation term, i.e., the cross-entropy with the labels plus $\mathcal{L}_{\mathrm{out}}(\cdot)$ for classification, and the squared error to the target plus $\mathcal{L}_{\mathrm{out}}(\cdot)$ for forecasting/regression. Algorithm~\ref{alg:conv} summarizes the model conversion process.

\begin{algorithm}[t]
\caption{Conversion of a trained Mixer model}\label{alg:conv}
\begin{algorithmic}[1]
\Require Trained model, number of devices $P$, $d$, $L$
\Ensure Split model: device ranges, sender codes, receiver tables
\State Split the layers into $P$ contiguous ranges; Mark the last layer of every sending device as a split layer
\State \textbf{Calibrate:} Collect rows $\mathbf{u}_t$ of every split layer on training data; Initialize $\mathbf{A},\mathbf{B}$ by principal components; Initialize $\mathbf{W}_4,\mathbf{b}_4$ to zeros, and $\mathbf{W}_3,\mathbf{b}_3$ from $U[-1/\sqrt{d},1/\sqrt{d}]$
\State \textbf{Step 1:} Freeze the original weights; Train the code parameters with $\mathcal{L}_1$~\eqref{eq:step1}; Keep the best model on validation
\State \textbf{Step 2:} Train all parameters with the task loss and $\mathcal{L}_{\mathrm{out}}$; Keep the best model on validation
\For{every split layer}
\If{$b\le b_{\max}$} \State \algparbox{Compute $\mathbf{Y}[\kappa]$ for $\kappa=0,\dots,K-1$ by~\eqref{eq:table}\\Store it on the receiving device}
\Else \State \algparbox{Store the decoder and the channel-mixing sublayer on the receiving device}
\EndIf
\EndFor
\end{algorithmic}
\end{algorithm}

\subsection{Tiny Time Mixers}\label{sec:ttm}
In TTM, every channel of a series is a separate patches-by-features table, and a mixing layer applies gated patch mixing followed by gated feature mixing. \method{} splits a mixing layer between these two sublayers, i.e., the sending device computes the patch mixing and codes every patch row of every channel, and the receiving device reads the feature-mixing output from its table. Since TTM normalizes every series by its own mean and standard deviation (which are needed by the forecast head), these two values ($64$ bits per channel) are transmitted together with the codes and are included in all message sizes for TTM. TTM reshapes its table between backbone levels, and therefore the number of patches at a split point depends on the position of the split.

\subsection{Communication Volume, Computation and Memory Costs}\label{sec:volume}
Table~\ref{tab:payload} lists the data transmitted per input and split point, i.e., per image for MLP-Mixer-B/16 and per channel of one series for TTM. The original model transmits the $ND$ activations of the split layer, i.e., $32ND$ bits in FP32 and $16ND$ bits in FP16. For INT8 and INT4, we divide every row by its largest absolute value before rounding, following the vector-wise quantization of~\cite{dettmers2022llmint8}, and transmit this scale as one FP32 value per row, which adds $N$ values and $32N$ bits. Thus, activation quantization reduces the number of bits per value, but it still transmits $ND$ values, and INT4 reduces the number of bits only by a factor of $7.9$ for MLP-Mixer-B/16. \method{} transmits $Nb$ bits, independently of the model width $D$. For $b\le b_{\max}$, every token is represented by one code, i.e., $N$ values, and for $b>b_{\max}$ the sending device transmits the $d$ levels of every token, i.e., $Nd$ values of $\log_2L$ bits. Relative to FP32 activations, this corresponds to a reduction factor of $32D/b$, i.e., $2{,}048$, $1{,}024$, $512$ and $205$ for MLP-Mixer-B/16 ($D=768$) with $b=12$, $24$, $48$ and $120$, respectively. Even with $b=120$, \method{} transmits $25.8\times$ fewer bits than INT4 activations.

For TTM, every channel of a series is a separate table, and the TTM column of Table~\ref{tab:payload} reports the data per channel of one series at a split point with $16$ patches and $96$ features, including the mean and the standard deviation of the series ($64$ bits). Since TTM reshapes its table between backbone levels, the number of patches depends on the split layer, e.g., mixing layer $5$ has $8$ patches with $192$ features, so that $12$-bit codes take $160$ instead of $256$ bits per channel, while FP32 activations take the same number of bits. A series with $n$ channels transmits $n$ times as much data. 

A pipeline over $P$ devices has $P-1$ split points, and the transmitted data add up over them. For example, with $P=3$ and $b=12$, each of the two split points of MLP-Mixer-B/16 transmits the $2{,}352$ bits, i.e., $4{,}704$ bits per image in total, compared with $9{,}633{,}792$ bits for FP32 activations.

\begin{table}[t]
\caption{Data transmitted per input and split point.}\label{tab:payload}
\centering
\setlength{\tabcolsep}{4pt}
\resizebox{\columnwidth}{!}{%
\begin{tabular}{@{}lcccc@{}}
\toprule
Message & Values & Bits & MLP-Mixer-B/16 & TTM\\
\midrule
FP32 activations & $ND$ & $32ND$ & $4{,}816{,}896$ & $49{,}216$\\
FP16 activations & $ND$ & $16ND$ & $2{,}408{,}448$ & $24{,}640$\\
INT8 activations & $ND+N$ & $8ND+32N$ & $1{,}210{,}496$ & $12{,}864$\\
INT4 activations & $ND+N$ & $4ND+32N$ & $608{,}384$ & $6{,}720$\\ \hline
\method{}, $b=12$ & $N$ & $Nb$ & $2{,}352$ & $256$\\
\method{}, $b=24$ & $12N$ & $Nb$ & $4{,}704$ & $448$\\
\method{}, $b=48$ & $24N$ & $Nb$ & $9{,}408$ & $832$\\
\method{}, $b=120$ & $60N$ & $Nb$ & $23{,}520$ & $1{,}984$\\
\bottomrule
\end{tabular}}
\end{table}

The computations of the split layer depend on the length of the codes. On the sending device, the split layer requires the multiply-adds of its token-mixing sublayer and of the projection. For MLP-Mixer-B/16 this corresponds to $115.6$ million per image, i.e., $11.1\%$ of a layer, for token-mixing and $0.9$ million per image for $d=6$, and $9.0$ million for $d=60$ for the projection. For $b\le b_{\max}$, the receiving device replaces the channel-mixing sublayer by one table lookup per token, which removes $924.8$ million multiply-adds per image and split point, i.e., $7.3\%$ of the computations of the model. For $b>b_{\max}$, the receiving device computes the decoder and the channel-mixing sublayer. Since the channel-mixing sublayer is also computed when the original model is split over the same devices, it only moves from the sending to the receiving device, and the additional computations are those of the projection and of the decoder, i.e., $1.13\%$ of the computations of the model ($143$ million multiply-adds per image for $d=60$). Regarding memory, the sending device stores the token-mixing weights ($0.61$~MB) and the projection ($18$~kB for $d=6$). For $b\le b_{\max}$, the receiving device stores the table ($12.6$~MB for $b=12$ and $201$~MB for $b=16$) instead of the channel-mixing weights ($18.9$~MB), and for $b>b_{\max}$ it stores the channel-mixing weights and the decoder ($2.7$~MB for $d=60$).

\section{Experiments}\label{sec:experiments}
We evaluate \method{} on image classification with MLP-Mixer-B/16 and on time-series forecasting with TTM. We first describe the experimental setup. We then evaluate the accuracy of \method{} for codes of $120$ down to $12$ bits per token, and compare it with the transmission of FP16, INT8 and INT4 activations, with and without fine-tuning.

\subsection{Experimental Setup}

For image classification we use MLP-Mixer-B/16~\cite{tolstikhin2021mixer} ($12$ layers, $N=196$, $D=768$, $D_{\mathrm{tok}}=384$, $D_{\mathrm{ch}}=3072$) at $224\times224$ pixels, initialized with the pre-trained weights of the timm model \texttt{mixer\_b16\_224}~\cite{wightman2019timm}. With $P=2$ devices, the split layer is layer $6$, and with $P=3$, the split layers are layers $4$ and $8$. For time-series forecasting we use TTM r2~\cite{ekambaram2024ttm,granitetsfm} with context length $512$ and horizon $96$, whose backbone and decoder contain $8$ mixing layers. With $P=2$, the split layer is mixing layer $4$, and with $P=3$, the split layers are mixing layers $3$ and $5$.
We refer to the \method{} versions of the two models as \methodmlp{} and \methodttm{}. 
Table~\ref{tab:data} summarizes the datasets. For CIFAR-10 and CIFAR-100~\cite{krizhevsky2009cifar}, a fixed $10\%$ of the training images is held out for validation. For ETTm1 and ETTm2~\cite{zhou2021informer}, we use the standard $12/4/4$-month split, and for Weather~\cite{wu2021autoformer} the standard $70/10/20\%$ split. Every channel is standardized with its training mean and standard deviation, and the MSE is computed on standardized data over all test windows.

\begin{table}[t]
\caption{Datasets information, with training/validation/test splits.}
\label{tab:data}
\centering
\footnotesize
\setlength{\tabcolsep}{3pt}
\begin{tabular}{@{}lll@{}}
\toprule
Dataset & Type & Train / val / test\\
\midrule
CIFAR-10 & 10 classes & $45{,}000$ / $5{,}000$ / $10{,}000$ images\\
CIFAR-100 & 100 classes & $45{,}000$ / $5{,}000$ / $10{,}000$ images\\
ETTm1, ETTm2 & 7 ch., 15 min & $34{,}560$ / $11{,}520$ / $11{,}520$ steps\\
Weather & 21 ch., 10 min & $70$ / $10$ / $20\%$ of the steps\\
\bottomrule
\end{tabular}
\end{table}

The original MLP-Mixer-B/16 models are fine-tuned from the pre-trained weights with SGD (momentum $0.9$, learning rate $0.01$, batch size $128$, mixed precision) for $40$ epochs with random-resized crops ($50$--$100\%$ of the image) and horizontal flips, and the model at the epoch with the best validation accuracy is kept. For TTM, the decoder and the forecast head of the pre-trained model are fine-tuned on the training windows (AdamW, learning rate $10^{-3}$, batches of $512$ windows, at most $1{,}000$ steps per epoch, $20$ epochs), and the model at the epoch with the best validation MSE is kept.

Unless stated otherwise, \method{} uses $L=4$ levels and $d=6$ dimensions ($b=12$ bits, $K=4{,}096$ codes), a linear encoder, and a decoder MLP with hidden width $768$, and the table is used for codes of up to $b_{\max}=16$ bits. The code is calibrated on $8$ training batches ($128$ images or $512$ series each). In the first step, the encoding parameters $\{\mathbf{A},\mathbf{a},\mathbf{B},\mathbf{b},g\}$ are trained for $40$ epochs with a learning rate of $10^{-3}$, and the temperature parameters $\vartheta_i$ with a learning rate of $10^{-2}$. In the second step, all parameters are trained for $40$ epochs, with learning rates of $5\times10^{-5}$ (MLP-Mixer) and $10^{-4}$ (TTM) for the weights of the original model, $10^{-4}$ for the encoding parameters, and $10^{-2}$ for the temperature parameters. Both steps use AdamW without weight decay, gradient clipping at norm $1$, and a cosine schedule (with a linear warm-up over the first $5\%$ of the steps for MLP-Mixer), and the model at the epoch with the best validation accuracy (or MSE) is kept. MLP-Mixer uses batches of $32$ images and a distillation temperature $T=2$, and TTM uses batches of $512$ series, i.e., (window, channel) pairs, with at most $1{,}000$ batches per epoch.

The reference is the original model split over the same devices, transmitting FP32 activations. We compare \method{} with the transmission of FP16, INT8 and INT4 activations (with the per-row absolute-maximum scaling of Section~\ref{sec:volume}) by the original model without fine-tuning. We also fine-tune the INT4 model end to end with a straight-through gradient (quantization-aware training~\cite{jacob2018quantization}), using the same loss as the second step of \method{} and the same total budget ($80$ epochs).

Every setting is run with three seeds, each with its own original model. We report the test accuracy (for MLP-Mixer) or the test MSE (for TTM) of the split model running from codes, as mean $\pm$ standard deviation over the seeds, together with its change against the original model of the same seed. The experiments ran on NVIDIA V100 GPUs.

\subsection{Accuracy Against the Bits per Token}\label{sec:curve}
\begin{figure*}[t]
\centering
\begin{tikzpicture}
\begin{axis}[name=left, width=0.47\textwidth, height=5.4cm, xmode=log, log basis x=2, xtick={12,24,48,120},
  xticklabels={12,24,48,120}, xlabel={bits per token $b$}, ylabel={$\Delta$ accuracy (points)}, ymin=-25, ymax=1,
  grid=major, legend style={font=\scriptsize, at={(0.5,-0.32)}, anchor=north, legend columns=2}, title={MLP-Mixer-B/16}, tick label style={font=\footnotesize}]
\addplot[blue, mark=*] coordinates {(12,-3.19) (24,-1.91) (48,-1.18) (120,-0.96)}; \addlegendentry{CIFAR-10, $P=2$}
\addplot[blue, dashed, mark=o] coordinates {(12,-7.68) (24,-5.84) (48,-3.41) (120,-2.06)}; \addlegendentry{CIFAR-10, $P=3$}
\addplot[red, mark=square*] coordinates {(12,-6.61) (24,-3.19) (48,-1.55) (120,-0.53)}; \addlegendentry{CIFAR-100, $P=2$}
\addplot[red, dashed, mark=square] coordinates {(12,-23.42) (24,-15.42) (48,-8.36) (120,-4.50)}; \addlegendentry{CIFAR-100, $P=3$}
\end{axis}
\begin{axis}[at={(left.east)}, anchor=west, xshift=1.6cm, width=0.47\textwidth, height=5.4cm, xmode=log, log basis x=2,
  xtick={12,24,48,120}, xticklabels={12,24,48,120}, xlabel={bits per token $b$}, ylabel={$\Delta$ MSE (\%)},
  ymin=-0.5, ymax=7, grid=major, legend style={font=\scriptsize, at={(0.5,-0.32)}, anchor=north, legend columns=3}, title={TTM}, tick label style={font=\footnotesize}]
\addplot[blue, mark=*] coordinates {(12,3.8) (24,2.4) (48,1.1) (120,0.5)}; \addlegendentry{ETTm1, $P=2$}
\addplot[blue, dashed, mark=o] coordinates {(12,6.3) (24,3.7) (48,2.5) (120,1.2)}; \addlegendentry{ETTm1, $P=3$}
\addplot[red, mark=square*] coordinates {(12,1.7) (24,1.2) (48,2.1) (120,1.3)}; \addlegendentry{ETTm2, $P=2$}
\addplot[red, dashed, mark=square] coordinates {(12,4.5) (24,2.1) (48,2.9) (120,1.8)}; \addlegendentry{ETTm2, $P=3$}
\addplot[black!60!green, mark=triangle*] coordinates {(12,0.4) (24,0.2) (48,0.5) (120,0.2)}; \addlegendentry{Weather, $P=2$}
\addplot[black!60!green, dashed, mark=triangle] coordinates {(12,1.5) (24,1.0) (48,0.5) (120,0.7)}; \addlegendentry{Weather, $P=3$}
\end{axis}
\end{tikzpicture}
\caption{Change against the original model as a function of the bits per token (mean over three seeds). Left: accuracy of MLP-Mixer-B/16. Right: test MSE of TTM (relative). At $12$ bits the receiving device reads a table, and above $b_{\max}=16$ bits it decodes the rows and computes the channel (feature) mixing.}
\label{fig:curve}
\end{figure*}
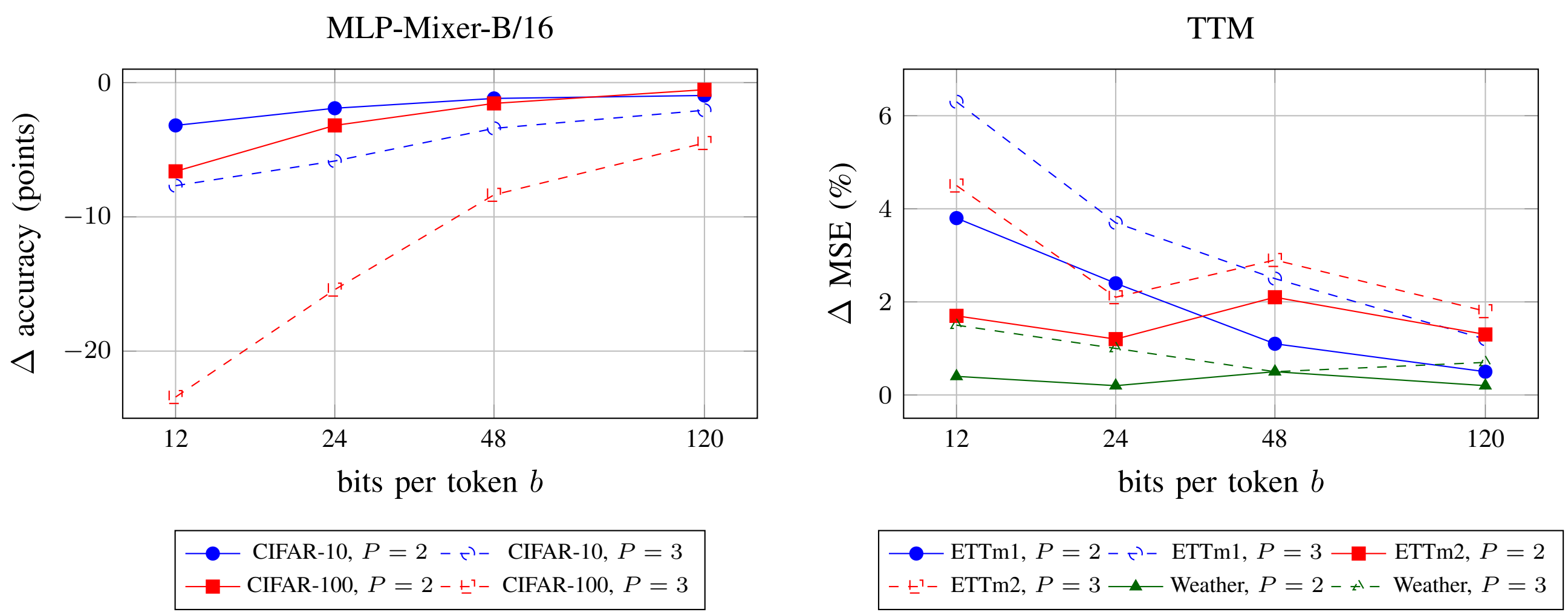

In this experiment, we evaluate how the accuracy of \method{} changes when the length of the codes is reduced. We report the results for $b=120$, $48$, $24$ and $12$ bits per token ($d=60$, $24$, $12$ and $6$) in Tables~\ref{tab:mixercurve} and~\ref{tab:ttmcurve} and in Fig.~\ref{fig:curve}. Changes in accuracy are given in percentage points, and changes in MSE relative to the MSE of the original model. Codes longer than $b_{\max}=16$ bits are decoded by the receiving device, as described in Section~\ref{sec:large}.

\begin{table}[t]
\caption{MLP-Mixer-B/16: test accuracy (\%) against bits per token $b$. Bytes per image over all split points, and reduction against FP32 activations. Mean $\pm$ standard deviation over three seeds.}
\label{tab:mixercurve}
\centering
\footnotesize
\setlength{\tabcolsep}{3pt}
\begin{tabular}{@{}llrrrcc@{}}
\toprule
Dataset & $P$ & $b$ & Bytes & Reduction & Accuracy & $\Delta$\\
\midrule
\multirow{4}{*}{CIFAR-10} & \multirow{4}{*}{2} & 12 & 294 & $2{,}048\times$ & $94.17\pm0.47$ & $-3.19\pm0.41$\\
 & & 24 & 588 & $1{,}024\times$ & $95.45\pm0.14$ & $-1.91\pm0.19$\\
 & & 48 & 1{,}176 & $512\times$ & $96.18\pm0.04$ & $-1.18\pm0.10$\\
 & & 120 & 2{,}940 & $205\times$ & $96.40\pm0.10$ & $-0.96\pm0.11$\\
\midrule
\multirow{4}{*}{CIFAR-10} & \multirow{4}{*}{3} & 12 & 588 & $2{,}048\times$ & $89.68\pm0.05$ & $-7.68\pm0.02$\\
 & & 24 & 1{,}176 & $1{,}024\times$ & $91.52\pm0.15$ & $-5.84\pm0.17$\\
 & & 48 & 2{,}352 & $512\times$ & $93.95\pm0.12$ & $-3.41\pm0.07$\\
 & & 120 & 5{,}880 & $205\times$ & $95.30\pm0.07$ & $-2.06\pm0.05$\\
\midrule
\multirow{4}{*}{CIFAR-100} & \multirow{4}{*}{2} & 12 & 294 & $2{,}048\times$ & $78.51\pm0.48$ & $-6.61\pm0.46$\\
 & & 24 & 588 & $1{,}024\times$ & $81.93\pm0.09$ & $-3.19\pm0.19$\\
 & & 48 & 1{,}176 & $512\times$ & $83.57\pm0.22$ & $-1.55\pm0.34$\\
 & & 120 & 2{,}940 & $205\times$ & $84.59\pm0.21$ & $-0.53\pm0.39$\\
\midrule
\multirow{4}{*}{CIFAR-100} & \multirow{4}{*}{3} & 12 & 588 & $2{,}048\times$ & $61.70\pm0.08$ & $-23.42\pm0.18$\\
 & & 24 & 1{,}176 & $1{,}024\times$ & $69.70\pm0.47$ & $-15.42\pm0.60$\\
 & & 48 & 2{,}352 & $512\times$ & $76.76\pm0.18$ & $-8.36\pm0.18$\\
 & & 120 & 5{,}880 & $205\times$ & $80.61\pm0.23$ & $-4.50\pm0.19$\\
\bottomrule
\end{tabular}
\end{table}

\begin{table}[t]
\caption{TTM: test MSE against bits per token $b$. Bits per channel of one series over all split points (including the mean and the standard deviation of the series), and reduction against FP32 activations. Mean $\pm$ standard deviation over three seeds.}\label{tab:ttmcurve}
\centering
\footnotesize
\setlength{\tabcolsep}{2.5pt}
\begin{tabular}{@{}llrrrcc@{}}
\toprule
Dataset & $P$ & $b$ & Bits & Reduction & MSE & $\Delta$\\
\midrule
\multirow{4}{*}{ETTm1} & \multirow{4}{*}{2} & 12 & 256 & $192.2\times$ & $0.2927\pm0.0012$ & $+3.8\%\pm0.7$\\
 & & 24 & 448 & $109.9\times$ & $0.2885\pm0.0008$ & $+2.4\%\pm0.2$\\
 & & 48 & 832 & $59.2\times$ & $0.2851\pm0.0002$ & $+1.1\%\pm0.4$\\
 & & 120 & 1{,}984 & $24.8\times$ & $0.2834\pm0.0006$ & $+0.5\%\pm0.6$\\
\midrule
\multirow{4}{*}{ETTm1} & \multirow{4}{*}{3} & 12 & 416 & $236.6\times$ & $0.2996\pm0.0010$ & $+6.3\%\pm0.8$\\
 & & 24 & 704 & $139.8\times$ & $0.2924\pm0.0014$ & $+3.7\%\pm0.8$\\
 & & 48 & 1{,}280 & $76.9\times$ & $0.2889\pm0.0010$ & $+2.5\%\pm0.8$\\
 & & 120 & 3{,}008 & $32.7\times$ & $0.2854\pm0.0010$ & $+1.2\%\pm0.7$\\
\midrule
\multirow{4}{*}{ETTm2} & \multirow{4}{*}{2} & 12 & 256 & $192.2\times$ & $0.1652\pm0.0005$ & $+1.7\%\pm0.4$\\
 & & 24 & 448 & $109.9\times$ & $0.1644\pm0.0007$ & $+1.2\%\pm0.5$\\
 & & 48 & 832 & $59.2\times$ & $0.1658\pm0.0008$ & $+2.1\%\pm0.7$\\
 & & 120 & 1{,}984 & $24.8\times$ & $0.1645\pm0.0006$ & $+1.3\%\pm0.3$\\
\midrule
\multirow{4}{*}{ETTm2} & \multirow{4}{*}{3} & 12 & 416 & $236.6\times$ & $0.1697\pm0.0001$ & $+4.5\%\pm0.2$\\
 & & 24 & 704 & $139.8\times$ & $0.1658\pm0.0005$ & $+2.1\%\pm0.2$\\
 & & 48 & 1{,}280 & $76.9\times$ & $0.1671\pm0.0005$ & $+2.9\%\pm0.4$\\
 & & 120 & 3{,}008 & $32.7\times$ & $0.1652\pm0.0011$ & $+1.8\%\pm0.5$\\
\midrule
\multirow{4}{*}{Weather} & \multirow{4}{*}{2} & 12 & 256 & $192.2\times$ & $0.1431\pm0.0004$ & $+0.4\%\pm0.0$\\
 & & 24 & 448 & $109.9\times$ & $0.1428\pm0.0002$ & $+0.2\%\pm0.4$\\
 & & 48 & 832 & $59.2\times$ & $0.1431\pm0.0005$ & $+0.5\%\pm0.4$\\
 & & 120 & 1{,}984 & $24.8\times$ & $0.1428\pm0.0001$ & $+0.2\%\pm0.3$\\
\midrule
\multirow{4}{*}{Weather} & \multirow{4}{*}{3} & 12 & 416 & $236.6\times$ & $0.1445\pm0.0004$ & $+1.5\%\pm0.6$\\
 & & 24 & 704 & $139.8\times$ & $0.1439\pm0.0001$ & $+1.0\%\pm0.4$\\
 & & 48 & 1{,}280 & $76.9\times$ & $0.1432\pm0.0005$ & $+0.5\%\pm0.6$\\
 & & 120 & 3{,}008 & $32.7\times$ & $0.1434\pm0.0003$ & $+0.7\%\pm0.5$\\
\bottomrule
\end{tabular}
\end{table}

With two devices and long codes, \method{} performs close to the original models, i.e., within ${\sim}2\%$ for codes of $48$ and $120$ bits per token, while it transmits two orders of magnitude fewer bits than FP32 activations for MLP-Mixer-B/16 and one to two orders of magnitude fewer bits per channel for TTM. Shorter codes increase the compression further with the accuracy decreasing gradually. For MLP-Mixer-B/16, the reduction in accuracy is larger on CIFAR-100 than on CIFAR-10, which suggests that a task with more classes needs more information per token. For TTM, the changes in MSE are small for all code lengths, and on Weather even the shortest codes lead to an MSE close to that of the original model. 
With three devices, the reduction in accuracy is larger, in particular for short codes and on CIFAR-100. In this case, the model is split twice, and the first split layer of MLP-Mixer-B/16 is layer $4$, i.e., earlier in the network.

For MLP-Mixer-B/16, the accuracy increases with the length of the codes, with diminishing returns, and longer codes help most in the settings with the largest losses. This indicates that the information carried by one code per token, rather than the training, limits the accuracy of short codes. For TTM, the MSE decreases consistently with the length of the codes on ETTm1, while on ETTm2 and Weather the differences between codes of $24$ bits or more are within the variation over the seeds.

\subsection{Comparison with Activation Quantization}
In this experiment, we compare \method{} with the transmission of quantized activations by the original model. Tables~\ref{tab:mixer} and~\ref{tab:ttm} report the results of FP16, INT8 and INT4 activations, together with the results of \method{} with $12$-bit codes, i.e., with the highest compression.

\begin{table*}[t]
\caption{MLP-Mixer-B/16: test accuracy (\%) of the original model and of \methodmlp{} with $12$ bits per token, and change against the original model ($\Delta$) of \methodmlp{} after each training step and of the original model with FP16, INT8 and INT4 activations, without and with fine-tuning (INT4-FT). Mean $\pm$ standard deviation over three seeds.}\label{tab:mixer}
\centering
\resizebox{\textwidth}{!}{%
\begin{tabular}{@{}llcccccccc@{}}
\toprule
Dataset & $P$ & Original & \method{} & $\Delta$ & $\Delta$ after step 1 & $\Delta$ FP16 & $\Delta$ INT8 & $\Delta$ INT4 & $\Delta$ INT4-FT\\
\midrule
CIFAR-10 & 2 & $97.36\pm0.07$ & $94.17\pm0.47$ & $-3.19\pm0.41$ & $-10.83\pm1.09$ & $-0.00\pm0.01$ & $-0.00\pm0.02$ & $-1.40\pm0.02$ & $+0.11\pm0.04$\\
CIFAR-10 & 3 & $97.36\pm0.07$ & $89.68\pm0.05$ & $-7.68\pm0.02$ & $-22.25\pm0.27$ & $-0.01\pm0.00$ & $+0.03\pm0.03$ & $-15.37\pm0.76$ & $+0.18\pm0.08$\\ \hline
CIFAR-100 & 2 & $85.12\pm0.18$ & $78.51\pm0.48$ & $-6.61\pm0.46$ & $-23.39\pm0.67$ & $0.00\pm0.00$ & $-0.04\pm0.02$ & $-6.89\pm0.21$ & $+2.06\pm0.14$\\
CIFAR-100 & 3 & $85.12\pm0.18$ & $61.70\pm0.08$ & $-23.42\pm0.18$ & $-46.24\pm0.60$ & $+0.01\pm0.03$ & $-0.12\pm0.11$ & $-41.98\pm3.50$ & $+1.95\pm0.11$\\
\midrule
\multicolumn{3}{@{}l}{Bits per image, $P=2$ / $P=3$} & $2{,}352$ / $4{,}704$ & & & $2{,}408{,}448$ / $4{,}816{,}896$ & $1{,}210{,}496$ / $2{,}420{,}992$ & $608{,}384$ / $1{,}216{,}768$ & $608{,}384$ / $1{,}216{,}768$\\
\bottomrule
\end{tabular}}
\end{table*}

\begin{table*}[t]
\caption{TTM: test MSE of the original model and of \methodttm{} with $12$ bits per token, and change against the original model ($\Delta$) of \methodttm{} after each training step and of the original model with FP16, INT8 and INT4 activations, without and with fine-tuning (INT4-FT). Mean $\pm$ standard deviation over three seeds.}\label{tab:ttm}
\centering
\resizebox{\textwidth}{!}{%
\begin{tabular}{@{}llcccccccc@{}}
\toprule
Dataset & $P$ & Original & \method{} & $\Delta$ & $\Delta$ after step 1 & $\Delta$ FP16 & $\Delta$ INT8 & $\Delta$ INT4 & $\Delta$ INT4-FT\\
\midrule
ETTm1 & 2 & $0.2819\pm0.0013$ & $0.2927\pm0.0012$ & $+3.8\%\pm0.7$ & $+11.5\%\pm0.3$ & $-0.0\%$ & $+0.0\%$ & $+0.5\%\pm0.0$ & $-0.4\%\pm0.4$\\
ETTm1 & 3 & $0.2819\pm0.0013$ & $0.2996\pm0.0010$ & $+6.3\%\pm0.8$ & $+18.3\%\pm0.1$ & $+0.0\%$ & $-0.0\%$ & $+0.6\%\pm0.2$ & $-0.5\%\pm0.2$\\ \hline
ETTm2 & 2 & $0.1624\pm0.0004$ & $0.1652\pm0.0005$ & $+1.7\%\pm0.4$ & $+6.2\%\pm0.6$ & $+0.0\%$ & $-0.0\%$ & $+0.2\%\pm0.0$ & $+1.5\%\pm0.6$\\
ETTm2 & 3 & $0.1624\pm0.0004$ & $0.1697\pm0.0001$ & $+4.5\%\pm0.2$ & $+10.7\%\pm0.4$ & $+0.0\%$ & $-0.0\%$ & $+0.7\%\pm0.0$ & $+1.2\%\pm0.5$\\ \hline
Weather & 2 & $0.1424\pm0.0004$ & $0.1431\pm0.0004$ & $+0.4\%\pm0.0$ & $+6.6\%\pm0.4$ & $-0.0\%$ & $+0.0\%$ & $+0.2\%\pm0.0$ & $+0.1\%\pm0.4$\\
Weather & 3 & $0.1424\pm0.0004$ & $0.1445\pm0.0004$ & $+1.5\%\pm0.6$ & $+10.7\%\pm0.1$ & $-0.0\%$ & $-0.0\%$ & $+1.7\%\pm0.1$ & $+0.2\%\pm0.4$\\
\midrule
\multicolumn{3}{@{}l}{Bits per channel, $P=2$ / $P=3$} & $256$ / $416$ & & & $24{,}640$ / $49{,}280$ & $12{,}864$ / $25{,}472$ & $6{,}720$ / $13{,}184$ & $6{,}720$ / $13{,}184$\\
\bottomrule
\end{tabular}}
\end{table*}

For MLP-Mixer-B/16, FP16 and INT8 activations do not change the accuracy, but they transmit about three orders of magnitude more bits than \methodmlp{} with $12$-bit codes. Post-training INT4 quantization still transmits more than two orders of magnitude more bits, and it reduces the accuracy considerably, in particular with three devices. It reduces the accuracy more than \methodmlp{} with $12$-bit codes in most settings, and more than \methodmlp{} with $120$-bit codes in all settings, although the latter transmits more than an order of magnitude fewer bits. Fine-tuned INT4 retains the original accuracy, and exceeds it on CIFAR-100 since its additional epochs of distillation also improve the network itself, but it transmits one to two orders of magnitude more bits than \methodmlp{}. Finally, the second training step of \methodmlp{} is essential, as the losses after the first step alone are several times larger.

TTM is robust to activation quantization, i.e., post-training INT4 changes the MSE only slightly, while it transmits more than an order of magnitude more bits per channel than \methodttm{} with $12$-bit codes. With $12$-bit codes, \methodttm{} increases the MSE more than post-training INT4 on ETTm1 and ETTm2, and the two perform similarly on Weather. With $120$-bit codes, which still transmit about three to four times fewer bits per channel than INT4, \methodttm{} is on par with or better than post-training INT4 in three of the six settings and slightly worse in the other three. Fine-tuning improves INT4 on ETTm1 and Weather, but it increases the MSE on ETTm2 probably due to overfitting.

\section{Conclusion}\label{sec:conclusion}
In this paper, we proposed \method{}, a framework for pipeline-distributed inference of Mixer models that transmits one short integer code per token instead of the complete intermediate activation. By placing every split point between the token-mixing and the channel-mixing sublayers of a Mixer layer, the sending device encodes every row of the token-mixed table without a search. The receiving device either reads the channel-mixing output from a table, which also absorbs the computations of a nonlinear decoder, or, for codes that are too long for its memory, decodes every row and computes the channel-mixing sublayer as an ordinary layer. A trained model is converted in two steps, i.e., the code is first learned with the network frozen and is then fine-tuned together with the network by distillation, and the split model computes the same output as the converted model. 
Our experiments with MLP-Mixer on image classification and with Tiny Time Mixers on time-series forecasting showed that the length of the codes controls a trade-off between compression and accuracy. With two devices, long codes keep the accuracy close to that of the original models while transmitting orders of magnitude less data than full-precision activations, and shorter codes increase the compression further with a gradual reduction in accuracy. The reduction in accuracy is larger for more difficult tasks and when the model is split over more devices. Compared with activation quantization, \method{} transmits considerably less data, while fine-tuned low-bit activations are generally more accurate at larger message sizes.

\section*{AI use statement}
In this work, generative AI tools were used to create and edit code, and as an additional way to find related work. All AI-assisted work was reviewed and edited as needed and the author takes full responsibility for the content of the paper.

\section*{Acknowledgment}
We acknowledge funding from the European Union’s Horizon Europe research and innovation actions under grant agreement No 101298393. The work only reflects the authors’ views. The EU Agency is not responsible for any use that may be made of the information it contains.

\bibliographystyle{IEEEtran}
\bibliography{refs}

\end{document}